\documentclass[runningheads]{llncs}
\usepackage{graphicx}

\usepackage{tikz}
\usepackage{comment}
\usepackage{amsmath,amssymb} 
\usepackage{color}

\usepackage[accsupp]{axessibility}  

\usepackage{epsfig}
\usepackage{graphicx}
\usepackage[utf8]{inputenc} 
\usepackage[T1]{fontenc}    
\usepackage{url}            
\usepackage{booktabs}       
\usepackage{amsfonts}       
\usepackage{nicefrac}       
\usepackage{microtype}      
\usepackage{amssymb}
\usepackage{amsmath}
\usepackage{bbm}
\usepackage{wrapfig}

\usepackage{algorithm}
\usepackage{subfigure}
\usepackage{epstopdf}
\usepackage{booktabs}
\usepackage{array}
\usepackage{multirow}
\usepackage{hhline}
\usepackage{makecell}
\usepackage{framed}
\usepackage{enumitem}
\usepackage{textcomp}
\usepackage[noend]{algpseudocode}
\usepackage{xcolor}

\makeatletter
\newcommand{\printfnsymbol}[1]{%
  \textsuperscript{\@fnsymbol{#1}}%
}
\makeatother

\makeatletter
\def\@fnsymbol#1{\ensuremath{\ifcase#1\or *\or \dagger\or \ddagger\or
  \mathsection\or \mathparagraph\or \|\or **\or \dagger\dagger
  \or \ddagger\ddagger \else\@ctrerr\fi}}
\makeatother

\usepackage[pagebackref,breaklinks,colorlinks]{hyperref}

\begin{document}
\pagestyle{headings}
\mainmatter
\def\ECCVSubNumber{5136}  

\graphicspath{{images/}}

\title{Towards Accurate Active Camera Localization}

\titlerunning{Towards Accurate Active Camera Localization}

\author{Qihang Fang$^{1}$\thanks{Equal contribution; ordered alphabetically.}, Yingda Yin$^{2}$\printfnsymbol{1}, Qingnan Fan$^3$\thanks{Corresponding authors.}, Fei Xia$^4$, Siyan Dong$^1$ \\Sheng Wang$^5$, Jue Wang$^3$, Leonidas Guibas$^4$, Baoquan Chen$^2$\printfnsymbol{2}}
\authorrunning{Q. Fang et al.}

\institute{$^{1}$Shandong University, $^{2}$Peking University\\
 $^{3}$Tencent AI Lab,  $^{4}$Stanford University,  $^{5}$3vjia \\
}

\maketitle

\begin{abstract}
In this work, we tackle the problem of active camera localization, which controls the camera movements actively to achieve an accurate camera pose. The past solutions are mostly based on Markov Localization, which reduces the position-wise camera uncertainty for localization. These approaches localize the camera in the discrete pose space and are agnostic to the localization-driven scene property, which restricts the camera pose accuracy in the coarse scale. We propose to overcome these limitations via a novel active camera localization algorithm, composed of a passive and an active localization module. The former optimizes the camera pose in the continuous pose space by establishing point-wise camera-world correspondences. The latter explicitly models the scene and camera uncertainty components to plan the right path for accurate camera pose estimation. We validate our algorithm on the challenging localization scenarios from both synthetic and scanned real-world indoor scenes. Experimental results demonstrate that our algorithm outperforms both the state-of-the-art Markov Localization based approach and other compared approaches on the fine-scale camera pose accuracy. Code and data are released at \href{https://github.com/qhFang/AccurateACL}{https://github.com/qhFang/AccurateACL}.
\end{abstract}

\section{Introduction}

The problem of camera localization is to estimate the accurate camera pose in a known environment. Such a problem is of great importance in many computer vision and robotics applications \cite{ye2022multi,zhong2019ad,zhong2021towards,luo2019end}. The research efforts in the past decades have been mostly devoted to camera localization in a passive manner \cite{shotton2013scene,meng2018exploiting,valentin2015exploiting,cavallari2017fly,cavallari2019real,brachmann2017dsac,brachmann2018learning,yang2019sanet}, which predicts the camera pose from the provided RGB/RGB-D frame. However, the passive localization approaches become unstable and fragile when they run into many well-known localization challenges, such as repetitive objects \cite{halber2019rescan} and textureless regions \cite{bui20206d}.

To resolve the aforementioned issues, the ability of active camera movement has been deployed in a set of works \cite{fox1998active,jensfelt2001active,mariottini2011active,chaplot2018active}, also known as \textit{active camera localization}. Three critical questions need to be answered to solve such a problem: 1) How to locate: how to localize the camera for the most accurate camera pose. 2) Where to go: the camera is initialized at an unknown position in the environment, where it should move for accurate active localization. As there are numerous localizable positions in the continuous camera pose space, the problem of active localization becomes highly ambiguous and difficult to solve. 3) When to stop: the agent is unaware of its ground truth camera pose, hence when it should decide to stop the camera movement.

Due to the difficulties raised by these questions, there has been very little research in this field. Most active localization works are inspired by Markov Localization \cite{cassandra1996acting}, a passive localization approach that takes random actions to reduce camera uncertainty within a 2D discrete belief map by Bayesian filtering. To decide camera movements, the early research of active localization \cite{fox1998active} handcrafts greedy heuristics to minimize the camera uncertainty in the coming step, while the recent work \cite{chaplot2018active} deploys a policy network to directly estimate the camera movement for higher localization accuracy via reinforcement learning. 
These approaches have dominated the active localization field in the past few decades. However, they still suffer from a few drawbacks that make them prohibitive for practical applications: 
1) \textit{Camera localization in the coarse-scale discrete pose space}. The localization accuracy relies on the predefined resolution of the 2D discrete belief map (40cm, 90$^{\circ}$ \cite{chaplot2018active}), which is usually unsatisfactory for many practical applications. Pursuing fine-scale accuracy (5cm, 5$^{\circ}$) would result in significantly increased state space, which is both memory and computation inefficient, and not scalable to large environments and continuous camera pose space.
2) \textit{Camera movement agnostic to localization-driven scene uncertainty}. The past approaches control the actions mainly based on the camera uncertainty, without considering the localization-driven scene uncertainty information much. Scene uncertainty is an intrinsic scene property, which is small for geometry- and texture- rich regions and large for repetitive and textureless regions (common localization challenges). Scene uncertainty serves as the important guidance for camera movements towards the localizable scene region, and ignorance of such information limits the localization accuracy.

To overcome the limitations exhibited in the existing approaches, we propose a novel active camera localization algorithm solved by reinforcement learning for accurate camera localization. Our algorithm consists of two functional modules, the \textit{passive localization} module and the \textit{active localization} module. 
The former passive module answers the ``How to locate'' question, and estimates the step-wise camera pose in the entire episode. It abandons localization in the discrete pose space, instead learns to predict the world coordinates from the single RGB-D frame, and optimizes the instant camera pose in the continuous pose space via the established camera-world coordinate correspondences. 
The latter active module consists of the scene uncertainty and camera uncertainty components that answer the ``Where to go'' and ``When to stop'' questions separately. 
The scene uncertainty component explicitly models the localization-driven scene properties and instant localization estimations in the scene, hence it aims to guide the camera movement towards the localizable region. 
The camera uncertainty component explicitly models the quality of camera pose estimations, and determines the adaptive stop condition for the camera movement.

We validate our algorithm on both the synthetic and scanned real-world indoor scenes.
Experimental results demonstrate that our proposed algorithm is able to achieve very high fine-scale camera pose accuracy (5cm, 5$^\circ$) compared to the Markov Localization based approach and other baselines. Benefited from the proposed scene uncertainty and camera uncertainty components, our algorithm learns various intelligent behaviors.

\section{Related Work}

\textbf{Passive localization.} The past camera localization approaches are mostly passive. They can be separated into two categories, which mainly differ in the input that comes from a single frame or a sequence of frames. 

For single-frame camera localization, one trend focuses on direct camera pose estimations by retrieving the most similar database image for the pose approximation of a reference image \cite{sattler2011fast,arandjelovic2016netvlad,sarlin2019coarse,taira2018inloc} or directly regressing the camera pose through neural networks \cite{kendall2015posenet,kendall2017geometric,wang2019atloc,brahmbhatt2018geometry}.
The other trend is indirect pose estimation that employs a two-step procedure, where the first step is to regress the 3D scene coordinates from the input RGB/RGB-D observation, and the second step takes a RANSAC based optimization to produce the final camera pose. The popular scene coordinate regression approaches are implemented as a decision tree \cite{shotton2013scene,meng2018exploiting,valentin2015exploiting,cavallari2017fly,cavallari2019real}
or a convolution neural network \cite{brachmann2017dsac,brachmann2018learning,yang2019sanet}.
These approaches builds structure-based knowledge in a more explicit way, and performs better than image retrieval on small- or middle- scale environments.

For temporal camera localization, one trend focuses on extending PoseNet to the time domain \cite{clark2017vidloc,valada2018deep,radwan2018vlocnet++,xue2019local}, whose performance is however limited by the image retrieval nature of PoseNet, as pointed out by \cite{sattler2019understanding}. The other more popular trend assumes a uniform belief of the current camera pose, and leverages Bayesian filtering to iteratively maximize the belief until a certain stop condition is reached. According to the representations of the belief, these approaches can be separated into Kalman Filter \cite{cox1994modeling,roumeliotis2000bayesian,zhou2020kfnet}, Markov Localization \cite{fox1998markov,fox1999markov} and Monte-Carlo localization \cite{thrun2001robust,dellaert1999monte}. Most active localization approaches are developed based on Markov Localization, which characterizes the belief as a 2D discrete map grid and the belief is maximized when the camera randomly navigates in the environment. However, Markov Localization suffers from expensive computation due to the huge state space for step-wise comparison.

\textbf{Active localization.} The pioneering work in active localization is active Markov Localization \cite{cassandra1996acting}, which adopts the greedy strategy for action selection to reduce the camera pose uncertainty. This work inspires a few followups \cite{jensfelt2001active,mariottini2011active}. However, as the problem of active localization is highly ambiguous, the traditional approaches mostly fall into shortsighted solutions. Thanks to the rapid development of reinforcement learning, active neural localization (ANL) \cite{chaplot2018active} firstly learns a policy model to seek a more accurate solution from visual observations. 
All the above approaches benefit from Markov Localization, yet also suffer from the limited discrete camera pose space and ignorance of scene-specific localization knowledge, as discussed in the Introduction session.

\textbf{Navigation.} Visual navigation \cite{savva2019habitat,anderson2018evaluation,chaplot2020object} aims at reaching \textit{a specific target} with the guidance of points (PointNav), images (ImageNav), semantics (ObjNav), etc. 
In contrast, active visual localization targets reaching an accurate camera pose. As there are \textit{numerous localizable poses}, the problem becomes highly ambiguous and more difficult to solve \cite{chaplot2018active}.
On the other hand, localization is an essential module in a navigation system, where active localization algorithms could be adopted in a navigation pipeline to further improve its performance.

\section{Approach}

\subsection{Task Setup}
Initializing the camera at an unknown position and orientation in an environment, the problem of \textit{active camera localization} is to control the camera movement actively towards a better place to obtain an accurate camera pose. Such a task provides us with two inputs. 1) A sequence of RGB-D frames along with the corresponding ground truth camera poses, denoted as $\{I^{(i)}_{\text{basis}}, C^{(i)}_{\text{basis}}\}_{i=1}^m$, where $m$ is the number of frames, following previous works \cite{shotton2013scene,meng2018exploiting,valentin2015exploiting,cavallari2017fly,cavallari2019real}. Such a posed RGB-D stream can be easily obtained by the SLAM system \cite{mur2017orb} with visual odometry and loop closure and roughly covers the scene. It provides the basis for both passive and active localization.
2) The instant RGB-D frame $I^{(t)}$ obtained during active localization.

The entire procedure of our framework is as follows. With the initial RGB-D frame $I^{(0)}$, the passive localization module estimates the current camera pose $\hat{C}^{(0)}$, and the active localization module estimates the next action for camera movement and then obtains a new RGB-D frame. Such a process is iterated until the active localization module decides to stop the movement, and the final camera pose is chosen as the estimated camera pose at the last step. 
The entire framework is shown in Figure \ref{figure:framework}.
We also refer readers to Algorithm 1 in the supplementary material for the entire procedure.

\begin{figure*}[t]
\centering
	\includegraphics[width=1\linewidth]{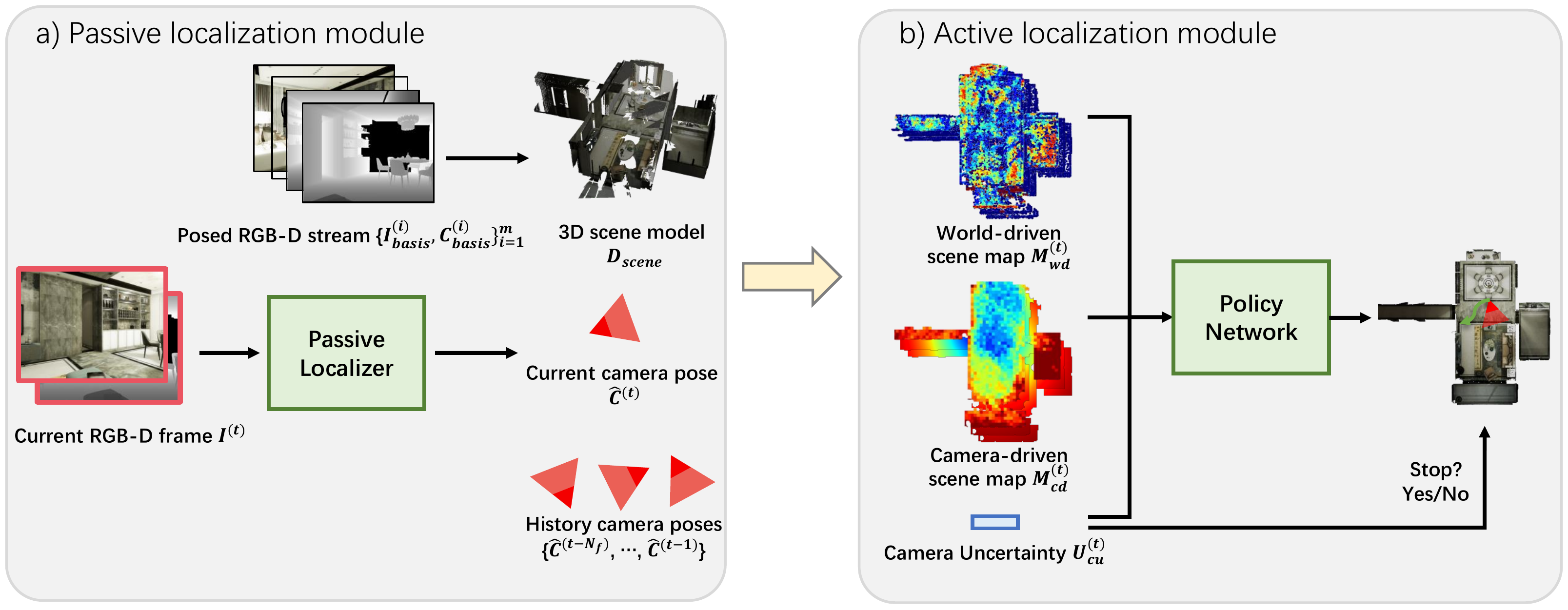}
	\caption{The full pipeline of our algorithm. a) Given the current RGB-D frame, the passive localizer estimates its camera pose, then b) the policy network takes the scene and camera uncertainty component to estimate the next action for camera movement, and the camera uncertainty component determines when to stop the movement. The 3D scene model is fused from the posed RGB-D stream, and further combined with the estimated current and history camera poses to construct the camera and scene uncertainty components.}
	\label{figure:framework}
\end{figure*}

\subsection{Passive Localization Module} \label{sec:passive}
The passive localization module answers the ``How to locate'' question. Instead of localizing the camera in the discrete pose space within a grid-based map as previous approaches \cite{fox1998active,chaplot2018active}, we propose to optimize the camera pose in the continuous pose space through a passive localizer. We adopt the state-of-the-art approach, decision tree \cite{cavallari2017fly}, to achieve this purpose thanks to its online adaption ability in novel scenes. We briefly describe it below\footnote[1]{Note we do not consider the implementation of passive localizer as our technical contribution, yet focus on how to make the best use of it for the entire task.}.

A decision tree, denoted as $DT$, takes a 2D image pixel $I^{(t)}_j$ sampled from the captured RGB-D frame $I^{(t)}$ as input, and performs hierarchical routing to estimate the index of one leaf node $DT(i)$, which consists of a set of 3D scene points $\{P_{dt,k}\}_{k \in \Omega_{dt,DT(i)}}$, where $\Omega_{dt,DT(i)}$ is the index set of 3D points belonging to the leaf node $DT(i)$ and $P_{dt,k}$ is back-projected in the world space with the posed RGB-D stream $\{I^{(t)}_{\text{basis}}, C^{(t)}_{\text{basis}}\}_{t=1}^m$. Then it randomly samples a 3D scene point from the distribution fitted from $\{P_{dt,k}\}_{k \in \Omega_{dt,DT(i)}}$ to establish the 2D-3D correspondence between the camera and world space. With correspondences obtained for many such input pixels, it infers the ranked camera pose hypotheses via pose optimization over the correspondences, and determines the camera pose $\hat{C}^{(t)}$ for the input frame $I^{(t)}$ by iteratively discarding the worse pose hypotheses until the last one left. 
The parameters of the decision tree lie in the split node determining the routing strategy. They are pre-trained on the 7-Scenes dataset \cite{shotton2013scene} and require no further finetuning. In novel scenes, only the leaf nodes are adaptively refilled online with the posed RGB-D stream\footnote[2]{Please refer to \cite{cavallari2017fly} for more implementation details of the decision tree.}.
The 3D scene model $D_{scene}$ is further constructed by fusing the posed RGB-D stream and the basis to generate the camera and scene uncertainty component for the active localization module.

\subsection{Active Localization Module}
In the vast literature of passive camera localization, two important factors have been studied widely for accurate localization. The first is \textit{camera uncertainty}, which indicates the confidence of camera pose estimations, and determines which camera pose to keep for localization \cite{shotton2013scene,cavallari2017fly,brachmann2019neural}.
The second is \textit{scene uncertainty}, which refers to the effectiveness of each scene region for accurate localization. For example, the passive localization approaches are able to achieve almost 100\% camera pose accuracy (5cm, 5$^\circ$) in scenes with small uncertainties, such as the texture- and geometry- rich scenes \cite{shotton2013scene,valentin2016learning}, yet underperform when there exhibit the scene regions with large uncertainties, such as textureless regions and repetitive objects \cite{bui20206d}, which are all the common localization challenges. We consider that both camera uncertainty and scene uncertainty are also necessary for accurate active localization, while the focus of most active localization works lies in camera uncertainty. Our active localization module consists of the scene uncertainty and camera uncertainty components, which answer the ``Where to go'' and ``When to stop'' questions separately.

\subsubsection{Scene Uncertainty Component}
Scene uncertainty is an intrinsic localization-driven scene property, and we describe such property from two perspectives, \textit{where the camera is located} and \textit{what underlying part of the scene is observed are more effective for accurate localization}. 
To model the above information, we propose the camera-driven scene map and world-driven scene map. They answer the ``Where to go'' question, and guide the camera movement towards scene regions with smaller uncertainties by combining the scene uncertainty property and the estimated camera properties (pose/world coordinate).
The scene uncertainty property is purely determined by the scene model $D_{scene}$ and the passive localization module, hence pre-computed and invariant to the active localization process, while the estimated camera properties are instantly computed from the captured RGB-D frame during the camera movements.

\begin{figure}[t]
	\centering
	\begin{tabular}{c}
 		\includegraphics[width=12cm]{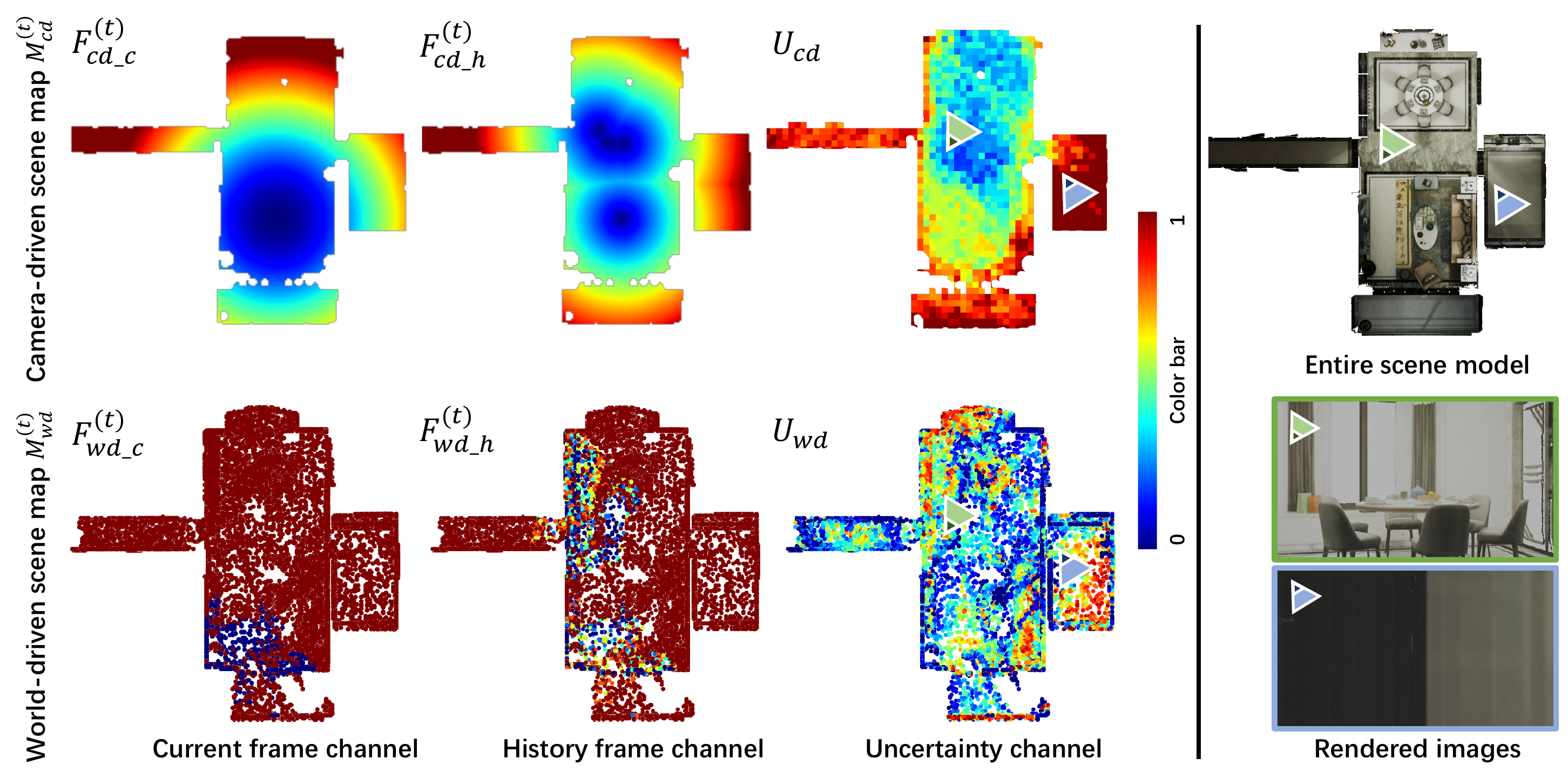}
	\end{tabular}
	\caption{Left: visualization of the different channels in both the camera-driven and world-driven scene maps. The value range of $F_{cd\_c}^{(t)}$ and $F_{cd\_h}^{(t)}$ is scaled into $[0,1]$ for better visualization with the color bar.
	Right: we render two first-view images with rich (green camera) or poor (blue camera) geometry and texture details, which are consistent with the uncertainty values shown in $U_{cd}$ and $U_{wd}$.
	}
	\label{figure:scene_uncertainty}
\end{figure}

\textbf{Camera-driven scene map: } 
The camera-driven scene map $M^{(t)}_{cd}$ at time step $t$ is represented in the form of the 2D top-view orthographic projection of the 3D scene model $D_{scene}$, and visualized in Figure \ref{figure:scene_uncertainty}. It consists of three components, position-wise uncertainty value $U_{cd}$, camera pose estimations of the current and history frames $F_{cd\_c}^{(t)},F_{cd\_h}^{(t)}$. The scene map $M^{(t)}_{cd}$ is computed as the position-wise concatenation of the three components and thus of size $X \times Y \times 3$, where $X$, $Y$ are the map size,
\begin{equation}
\small
M^{(t)}_{cd} = \text{Concat}\{U_{cd}, F_{cd\_c}^{(t)},F_{cd\_h}^{(t)}\}
\end{equation}
To filter out the invalid camera positions, we initialize all the map channels as the binary traversable map where the traversable and obstacle positions are filled with $0$ and $-1$ separately, and only update the values at traversable positions.

The uncertainty channel $U_{cd}$ describes the probability of successful passive localization at each valid camera position in the scene map.
To be specific, for each valid camera position, we render RGB-D frames along $N_{cd}$ uniformly sampled camera directions with the scene model $D_{scene}$, and estimate the corresponding camera poses via the passive localization module. The position-wise uncertainty value $U_{cd,i}$ is inversely proportional to the camera pose accuracy (within $\lambda_{cd}$ cm, $\lambda_{cd}$ degrees) averaged over all the rendered RGB-D frames,
\begin{equation}
\small
U_{cd,i} =  1 - \frac{1}{N_{cd}} \sum_{j \in [1, N_{cd}]} A^{(j)}
\end{equation}
where $A^{(j)}$ is the binary camera pose accuracy for the $j$th frame.

The current camera pose estimation channel $F_{cd\_c}^{(t)}$ indicates where the camera is located in the scene map estimated from the current RGB-D frame $I^{(t)}$. 
As the camera pose is estimated in the orientation-aware continuous space, and not compatible with the orientation-agnostic discrete scene map, to minimize this gap, we simply discretize the camera pose and project it onto the 2D scene map by only considering its translation on the horizontal plane. 
However, the estimated camera pose formulated in this way is nothing but a single point shown in the scene map, and tends to be overwhelmed by its blank neighborhood via the common convolution operations. To highlight the importance of the camera pose information in the 2D map, we draw a distance map centered on the discretized camera position via distance transform \cite{borgefors1986distance} as $F_{cd\_c}^{(t)}$. For the history camera pose estimation channel, we obtain the estimated camera positions in the 2D scene map for the last $N_f$ frames ($I^{(t-N_f)}, ..., I^{(t-1)}$) same as the current channel, and draw a distance map centered on the history camera positions via distance transform as $F_{cd\_h}^{(t)}$.

\setlength{\tabcolsep}{2pt}	

\textbf{World-driven scene map: } 
The world-driven scene map $M^{(t)}_{wd}$ at time step $t$ is represented in the form of the 3D point cloud sampled from the scene model $D_{scene}$, and visualized in Figure \ref{figure:scene_uncertainty} from the top view for better comparison with the camera-driven scene map. It consists of four components, the $x$, $y$, $z$ world coordinates of the scene points $P_{wd}$, point-wise uncertainty value $U_{wd}$, world coordinate estimations of the current and history frames $F^{(t)}_{wd\_c}, F^{(t)}_{wd\_h}$. The scene map $M^{(t)}_{wd}$ is computed as the point-wise concatenation of the four components and thus of size $N_{wd\_p} \times 6$ (with $N_{wd\_p}$ points and 6 channels),
\begin{equation}
\small
M^{(t)}_{wd} = \text{Concat}\{P_{wd}, U_{wd}, F^{(t)}_{wd\_c}, F^{(t)}_{wd\_h}\}
\end{equation}

The uncertainty channel $U_{wd}$ describes the effectiveness of each observable scene point to the successful passive localization, and the point-wise uncertainty value is highly related to the viewpoint where the scene point is observed. To compute the uncertainty value, we first render $N_{wd\_r}$ RGB-D frames that are randomly positioned and oriented within the traversable region. We associate each 3D scene point $P_{wd,i}$ with an index set of 2D image pixels $\Omega_{wd,i}$ that can be back-projected to it as follows,
\begin{equation}
\small
\Omega_{wd,i} = \{j| \forall j \in \Omega_{wd\_r}, \|P_{wd\_{r,j}}-P_{wd,i}\| < \lambda_{wd}\}
\end{equation}
where $\Omega_{wd\_r}$ is the index set of all the image pixels in the $N_{wd\_r}$ rendered frames, $P_{wd\_{r,j}}$ is the 3D point in the world space back-projected from the pixel $j$ in $\Omega_{wd\_r}$, and $\lambda_{wd}$ is a threshold and measured in centimeters.

Then for each 2D pixel, we evaluate its uncertainty value $U_{wd\_{r,j}}$ as the estimation quality of the passive localizer, which in our case is the routing quality of the decision tree and adapted from the common measurement for the camera pose evaluation \cite{shotton2013scene,cavallari2019real}. To be specific, $U_{wd\_{r,j}}$ is computed as a binary value that judges if its back-projected 3D point $P_{wd\_{r,j}}$ is close to any 3D point in its routed leaf node of the decision tree, 
\begin{equation}
\small
U_{wd\_{r,j}} = \begin{cases}
      0 & \text{$(\min_{k \in \Omega_{dt,DT(j)}} \|P_{wd\_{r,j}}-P_{dt,k}\|) < \lambda_{wd}$} \\
      1 & \text{otherwise}
    \end{cases}
\end{equation}
where $\Omega_{dt,DT(j)}$ is the index set of the 3D points $P_{dt,k}$ in the leaf node $DT(j)$ where the pixel $j$ is routed.
Then the uncertainty value of each 3D scene point $U_{wd,i}$ is averaged over the ones of its associated 2D pixels, 
\begin{equation}
\small
U_{wd,i} = \frac{1}{N_{wd,i}} \sum_{j \in \Omega_{wd,i}} U_{wd\_{r,j}}
\end{equation}
where $N_{wd,i}$ is the size of the index set $\Omega_{wd,i}$.

The current world coordinate estimation channel indicates where the world coordinates back-projected from the current RGB-D frame using the estimated camera pose are located on the scene point cloud, hence is computed as the point-wise binary value that describes if each scene point is occupied by at least one back-projected world coordinates. 
To be specific, for each scene point $P_{wd,i}$, its binary value $F^{(t)}_{wd\_c,i}$ is outputted by an indicator function based on the unidirectional Chamfer distance from the estimated world coordinates to the scene point,
\begin{equation}
\small
F^{(t)}_{wd\_c,i} = \begin{cases}
      0 & \text{$(\min_{l \in \Omega_{f}^{(t)}} \|P_{wd,i} - P_{f,l}^{(t)}\|_2^2) < \lambda_{wd}$} \\
      1 & \text{otherwise}
    \end{cases}
\end{equation}
where $\Omega_{f}^{(t)}$ is the index set of 3D points $P_{f,l}^{(t)}$ back-projected from the current frame $I^{(t)}$ with the estimated camera pose $\hat{C}^{(t)}$.

The history world coordinate estimation channel is simply averaged over the last $N_f$ frames. Specifically, $F^{(t)}_{wd\_h,i}$ is computed as,
\begin{equation}
\small
F^{(t)}_{wd\_h,i} = \frac{1}{N_f} \sum_{t' \in [1, N_f]} F^{(t-t')}_{wd\_c,i}
\end{equation}

\textbf{Analysis of scene uncertainty:}
We visualize the computed uncertainty channel in both the camera-driven and world-driven scene maps in Figure \ref{figure:scene_uncertainty}. The uncertainty value denotes how much the valid camera positions and observable scene points are uncertain to successful camera localization. For a better understanding of the computed uncertainty values, we also render two first-view images with the green and blue cameras separately in the scene. The blue camera captures an image with poor texture and geometry, which is a common localization challenge, correspondingly, its camera position and observed scene points in the uncertainty channel all contain very large uncertainties. On the other hand, The green camera observes an image with rich texture and geometry, which is usually easy for accurate localization, correspondingly, its camera position and observed scene points mostly contain small uncertainties. The above observation further validates the design of the proposed scene uncertainty component.

\subsubsection{Camera Uncertainty Component}
Camera uncertainty is an intrinsic camera property, which represents the quality of the current camera pose estimation during camera movements. The camera uncertainty component answers the ``When to stop'' question, and hence determines the adaptive stop condition for active movements. 
Ideally, the camera uncertainty value should be computed by directly comparing the estimated camera pose with the ground truth camera pose, which is however absent during active movements. To alleviate the above difficulty, instead of directly dealing with the camera pose, we propose to calculate the camera uncertainty value by comparing the captured depth observation that represents the ground truth camera pose and the depth image projected from the 3D scene model $D_{scene}$ with the estimated camera pose $\hat{C}^{(t)}$. To be specific, given the observed depth and projected depth images, we first back-project the two images into the point clouds in the camera space with the known camera intrinsic parameters. Then we leverage the recent colored point cloud registration approach (Colored ICP) \cite{park2017colored} to register the two point clouds and estimate the relative camera pose between them. When the two point clouds are roughly aligned, the adopted ICP approach is able to achieve very tight point cloud alignment. 
Therefore, the estimated relative pose indicates how far the current camera pose estimation $\hat{C}^{(t)}$ is to the ground truth, and is treated as the camera uncertainty component $U_{cu}^{(t)} \in \mathbb{R}^2$.

To ease policy learning, many previous works fix the episode length \cite{cassandra1996acting,jensfelt2001active,chaplot2018active} for camera movements, which is inefficient in implementation. In this work, we propose to adaptively stop the camera movement based on the proposed camera uncertainty component. To be specific, we consider a successful localization to stop the camera movement when the camera uncertainty component is within $\lambda_{cu}$ cm, $\lambda_{cu}$ degrees.

\begin{wrapfigure}[14]{r}{0.4\textwidth}
\begin{center}
    \includegraphics[width=0.43\textwidth]{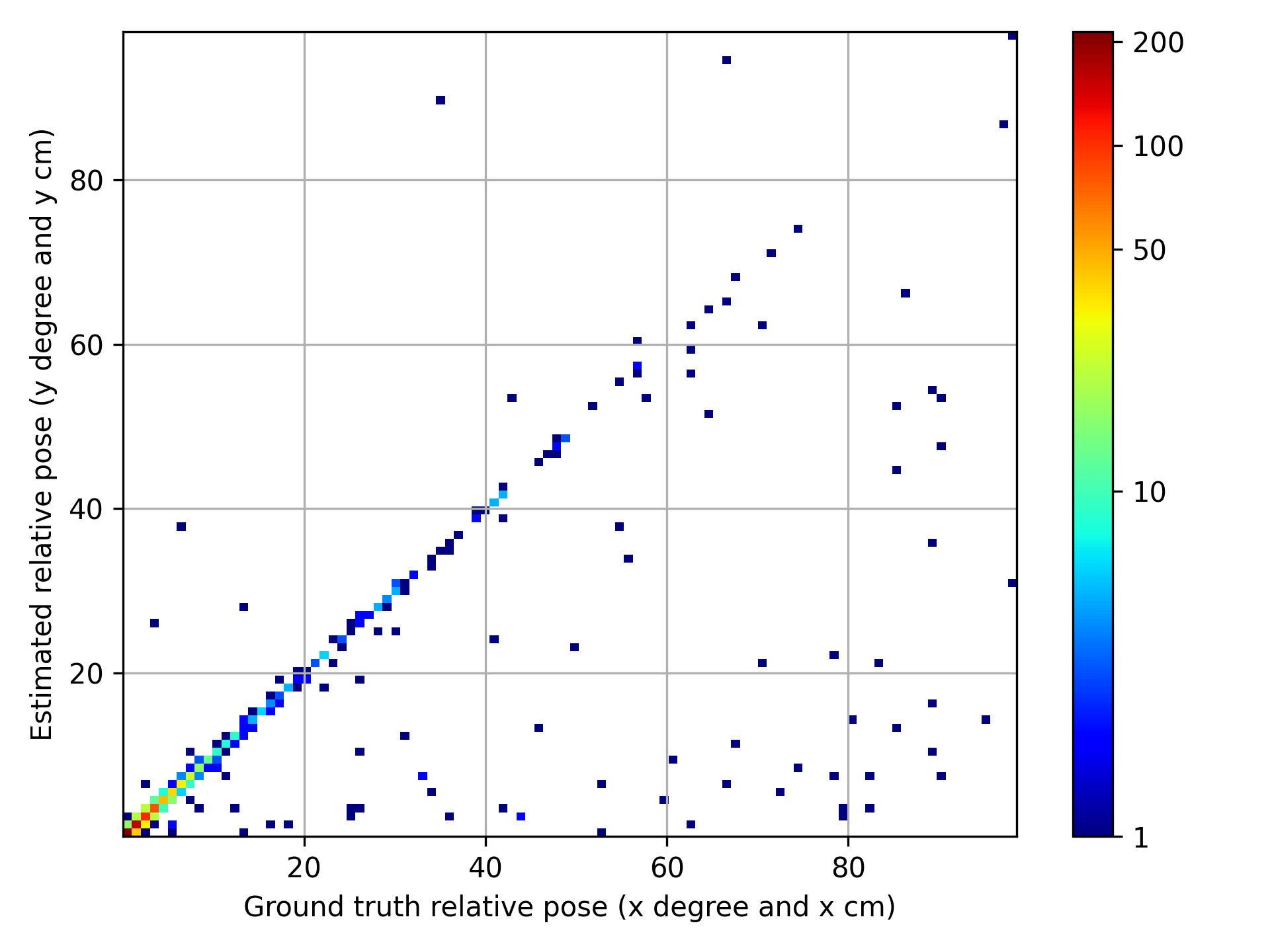}
\end{center}
	\caption{Justification of the camera uncertainty component. The color bar indicates the sample number.}
	\label{figure:camera_uncertainty}
\end{wrapfigure}

\textbf{Analysis of camera uncertainty:}
To justify the effectiveness of the camera uncertainty component, we evaluate how close the estimated relative pose is to the ground truth in Figure \ref{figure:camera_uncertainty}, which contains 4500 samples randomly collected in the indoor scenes introduced in Section \ref{sec:exp}. We can observe that most samples lie on diagonal lines, which means the relative pose estimations are accurate in general. 
To be specific, when the estimated relative poses are within 5cm, 5$^{\circ}$ (2362 samples), most samples (94.14\% = 2362/2509) are truly within 5cm, 5$^{\circ}$ compared to the ground truth (2509 samples). It means the adaptive stop condition judged by the camera uncertainty component is trustworthy.

\subsubsection{Reinforcement Learning Formulation}
We optimize the policy with the off-policy learning method Proximal Policy Optimization (PPO) \cite{schulman2017proximal} by maximizing the accumulated reward in the entire episode. The policy network is detailed in the supplementary material.

\textbf{Reward function:}
We design the reward $\mathcal{R}$, consisting of a slack reward $\mathcal{R}_{s}$ and an exploration reward $\mathcal{R}_{e}$. The slack reward punishes unnecessary steps and is defined as $\mathcal{R}_{s} = -0.1$, which gives a negative reward for every action performed. The exploration reward $\mathcal{R}_{e}$ awards the agent for visiting the unseen cells to avoid repeated traversal among the same region following \cite{ye2021auxiliary,ramakrishnan2021exploration}.
To achieve this, we maintain a 2D occupancy map with the same map size as the camera-driven scene map, and each cell is filled with the visit count from the episode initialization. Then $\mathcal{R}_{e} = 0.1/v$, where $v$ is the visit count in the currently occupied cell, whose position is obtained from the ground truth as the reward is only employed during training.
The final reward is the summation of both rewards, ${\mathcal{R}} = {\mathcal{R}}_{s}+{\mathcal{R}}_{e}$.

\textbf{Policy input:} 
The input of the policy should encode the knowledge of the sensor input and the scene, and have positive guidance for the agents to move towards more localizable regions acknowledged by the passive localization module. In order to achieve this goal, the policy takes the scene uncertainty and camera uncertainty components at time step $t$ as input $\{M^{(t)}_{cd}, M^{(t)}_{wd}, U_{cu}^{(t)}\}$.

\textbf{Action space:} 
Following the previous active localization setting \cite{fox1998active,chaplot2018active}, we assume that the agent (camera) moves with the 3-DoF (Degree of Freedom) action space within the 1-meter high 2D plane parallel to the ground. The agent is capable of performing three actions, \textit{move forward}, \textit{turn left} and \textit{turn right}. The agent moves forward by 20cm, and turns left/right by 30$^\circ$. We further disturb the actions with Gaussian noises as introduced in the supplementary material.

\section{Experiments}
\subsection{Experimental Setup}\label{sec:exp}

\textbf{Data processing:}
We evaluate our algorithm on both the synthetic and scanned real-world indoor scenes. To alleviate the difficulty of creating the common localization challenges in the synthetic data, we collect 35 high-quality indoor scenes with an average area of 40.9$m^2$, that feature textureless walls, repetitive pillows/drawings, \textit{etc}, by design, and provide a train/test split of the scenes (train/test: 15/20 scenes). For the scanned real-world data, we collect 5 indoor scenes with an average area of 64.8$m^2$ from the public Matterpot3D dataset \cite{chang2017matterport3d} only for evaluation. 
For each indoor scene, we provide a list of data as follows:
\begin{itemize}[leftmargin=0.4cm,nosep]
    \item A sequence of <RGB-D image, camera pose> pairs $\{I^{(t)}_{\text{basis}}, C^{(t)}_{\text{basis}}\}_{t=1}^m$ that provides the basis for localization and roughly covers the scene. 
    \item Instant RGB-D frame $I^{(t)}$ obtained during active localization.
    \item 100 test images in each test scene. They are randomly sampled in the scene region of large uncertainties to increase the localization difficulty (1 meter away from the positions of $U_{cd,i}\leq0.5$).
\end{itemize}

We name the synthetic dataset ACL-synthetic, and the real-world dataset ACL-real. Our algorithm is trained only on the train split of the ACL-synthetic dataset, and evaluated on both the test split of the ACL-synthetic dataset and the entire ACL-real dataset. 
During training, the camera is initialized randomly in the scene. During evaluation, the camera is initialized with one of the 100 test images.
More details about both datasets\footnote[3]{Note we do not claim the contribution of the collected indoor scenes, which can be replaced with any ones in public indoor scene datasets.} are in the supplementary material.

\textbf{Training setting:}
The passive localizer is adapted online in novel scenes with the posed RGB-D stream as mentioned in Section \ref{sec:passive}, hence only the policy network needs to be trained in our algorithm. 
Following the popular camera pose accuracy measured by 5cm, 5$^{\circ}$ \cite{shotton2013scene,meng2018exploiting,valentin2015exploiting,cavallari2017fly,cavallari2019real,brachmann2017dsac,brachmann2018learning,yang2019sanet}, we set $\lambda_{cd} = \lambda_{wd} = \lambda_{cu} = 5$. It means we encourage the agent to move to the scene region where the camera pose estimated from the passive localization module is within 5cm, 5$^{\circ}$ to the ground truth, and stop the camera movement when it believes the estimated camera pose is within 5cm, 5$^{\circ}$ to the ground truth.

\textbf{Evaluation metrics:}
The major goal of active camera localization lies in achieving higher camera pose accuracy. 
We evaluate the accuracy (\%) as the proportion of successful localization episodes whose translation and rotation error for the final camera pose is within 5cm, 5$^{\circ}$ by default, a fine-scale measurement compared to 40cm, 90$^{\circ}$ adopted in ANL \cite{chaplot2018active}. We further compute the number of steps (\#steps) taken to finish the successful localization acknowledged by the accuracy measure. It is only a complementary metric, while we value the accuracy most.
We limit all the approaches with a maximum step length of 100.

\subsection{Compared Approaches}

We detail the compared approaches below.
\begin{itemize}[leftmargin=0.4cm,nosep]
\item \textbf{No-movement}. It only uses the passive localization module to estimate the camera pose for the initial test frame. We adopt two passive localizers for comparison, the default decision tree \cite{cavallari2017fly} (No-movement (DecisionTree)) and the popular CNN-based passive localizer \cite{brachmann2017dsac} (No-movement (DSAC)).

\item \textbf{Turn-around}. This baseline works by turning a circle along the vertical axis for 12 uniformly-sampled directions without any forward movement, and stopping at the camera pose with the smallest camera uncertainty value.

\item \textbf{Camera-descent}. It iterates over all the possible actions in the future steps and selects the one with the smallest camera uncertainty value as the following path, hence it moves along the camera uncertainty descent direction. It stops when it triggers our adaptive stop condition. Depending on the number of explored future steps (1/2 steps), we derive two baselines, Camera-descent (t+1/t+2). We adopt beam search to implement Camera-descent (t+2) for memory efficiency.

\item \textbf{Scene-descent}. It assumes the estimated camera pose is roughly correct, and computes the shortest path from the estimated camera pose to the more localizable region ($U_{cd,i} \leq 0.5$) in the camera-driven uncertainty channel. Therefore, it moves along the scene uncertainty descent direction. It stops when it finishes the traversal over the shortest path.

\item \textbf{ANL}. 
Active neural localization (ANL) \cite{chaplot2018active} is a state-of-the-art active localization approach derived from the Markov localization. 
Due to the significant requirement of memory and computation resources, its camera pose is limited at the resolution of 20cm, 90$^{\circ}$ with Nvidia Tesla V100 of 32G memory in our implementation (40cm, 90$^{\circ}$ in \cite{chaplot2018active}).

\end{itemize}

\setlength{\tabcolsep}{10pt}
\begin{table}[t]
\centering
    \scriptsize
    \caption{Numerical results evaluated with the fine-scale 5cm, 5$^\circ$ accuracy.}
    \begin{tabular}{l|c|c|c|c}
        \toprule
        & \multicolumn{2}{c|}{ACL-synthetic} & \multicolumn{2}{c}{ACL-real} \\
        \cmidrule{1-5}
        Methods & Acc (\%) & \#steps & Acc (\%) & \#steps \\
        \midrule
        ANL \cite{chaplot2018active} & 3.25 & 100 & 3.20 & 100 \\
        No-movement (DecisionTree) & {9.35} & 0 & 6.80 & 0\\
        No-movement (DSAC) & {14.90} & 0 & 7.80 & 0\\
        Turn-around & {25.00} & 12 & 35.20 & 12\\
        Camera-descent (t+1) & 61.55 & 22.90 & 61.40 & 26.85\\
        Camera-descent (t+2) & 55.30 & 22.60 & {59.20} & {25.78}\\
        Scene-descent & 57.65 & 18.56 & 54.20 & 16.87\\
        \midrule
        Ours (w/o $\mathcal{R}_{e} \& M^{(t)}_{cd}$) & {67.65} & {17.40} & 70.60 & 19.71\\
        Ours (w/o $\mathcal{R}_{e} \& M^{(t)}_{wd}$) & {66.40} & {16.27} & 67.40 & 18.63\\
        Ours (w/o $\mathcal{R}_{e}$) & {72.50} & {18.57} & 73.00 & 20.72\\
        \midrule
        Ours & \textbf{83.05} & {17.33} & \textbf{82.40} & {17.90}\\
        \bottomrule
    \end{tabular}
\label{table:comparison_fine}
\end{table}

\setlength{\tabcolsep}{12pt}
\begin{table}[t]
    \centering
    \scriptsize
    \caption{Numerical results evaluated with the coarse-scale 20cm, 90$^\circ$ accuracy.}
    \begin{tabular}{l|c|c|c|c}
        \toprule
          & \multicolumn{2}{c|}{ACL-synthetic} & \multicolumn{2}{c}{ACL-real} \\
        \midrule
        Methods & Acc (\%) & \#steps & Acc (\%) & \#steps \\
        \midrule
        Markov Loc. \cite{fox1998markov} & 44.70 & 100  & 39.20  & 100   \\
        Active Markov Loc. \cite{fox1998active} & 44.10 &  100 & 40.00 & 100 \\
        ANL \cite{chaplot2018active} & 87.30 & 100 & 84.20 & 100  \\
        \midrule
        Ours  & \textbf{88.75} & {17.09} & \textbf{85.20} & {17.88}  \\
        \bottomrule
    \end{tabular}
    \label{table:comparison_coarse}
\end{table}

\subsection{Results}

\begin{figure*}[t]
\centering
    \includegraphics[height=3.6cm]{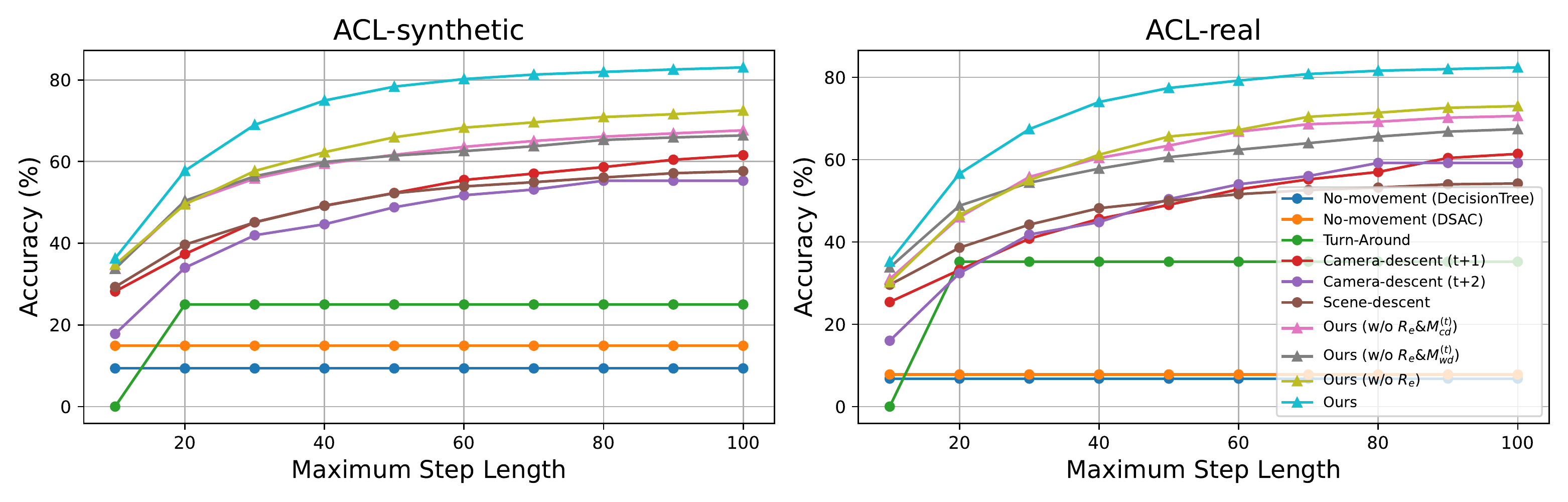}
	\caption{Plot of the localization accuracy that varies with different maximum step lengths.}
	\label{figure:curve}
\end{figure*}

\textbf{Comparison with baselines:}
The comparison is shown in Table \ref{table:comparison_fine}.
We analyze the results in the synthetic indoor scenes (ACL-synthetic) first.
The No-movement baselines achieve upmost 14.90\% accuracy, indicating the fact that passive localization is not sufficient in our challenging localization scenarios. By enabling the rotation actions, the accuracy of the Turn-around heuristic is only 25.00\% at most, which suggests the importance of active camera movements. The Camera-descent and Scene-descent baselines contain smarter designs based on our proposed camera uncertainty and scene uncertainty components, and also significantly improve the accuracy. 
Our algorithm outperforms all the approaches in the camera pose accuracy (83.05\%) with limited steps being taken. Similar phenomenons can also be observed in the scanned real-world indoor scenes (ACL-real).
We further visualize the accuracy that progresses along the increasing maximum step length in Figure \ref{figure:curve}, where our algorithm is consistently better than all the others.

\textbf{Comparison with ANL:} 
ANL is trained on the discrete belief map of resolution 20cm, 90$^{\circ}$, which is almost the upper bound of the camera pose scale it can achieve. Therefore, it performs poorly on the finer-scale accuracy (5cm, 5$^{\circ}$) as expected.
Furthermore, we evaluate both ANL and our method on the coarse-scale 20cm, 90$^\circ$ accuracy where ANL is good at, shown in Table \ref{table:comparison_coarse}. We can see that ANL achieves significantly better results on the coarser-scale accuracy, while our method still achieves comparable localization accuracy, with much fewer moving steps. We also compare with Markov localization \cite{fox1998markov} and  Active Markov Localization \cite{fox1998active} on the coarse-scale 20cm, 90$^\circ$ accuracy following \cite{chaplot2018active}.

\textbf{Ablation study:} We justify our algorithm by ablating three components, the exploration reward $\mathcal{R}_{e}$, camera-driven scene map $M^{(t)}_{cd}$ and world-driven scene map $M^{(t)}_{wd}$. Experimentally, we observe that our algorithm benefits from all three components.

\begin{figure}[t]
	\centering
	\begin{tabular}{c}
 		\includegraphics[width=11cm]{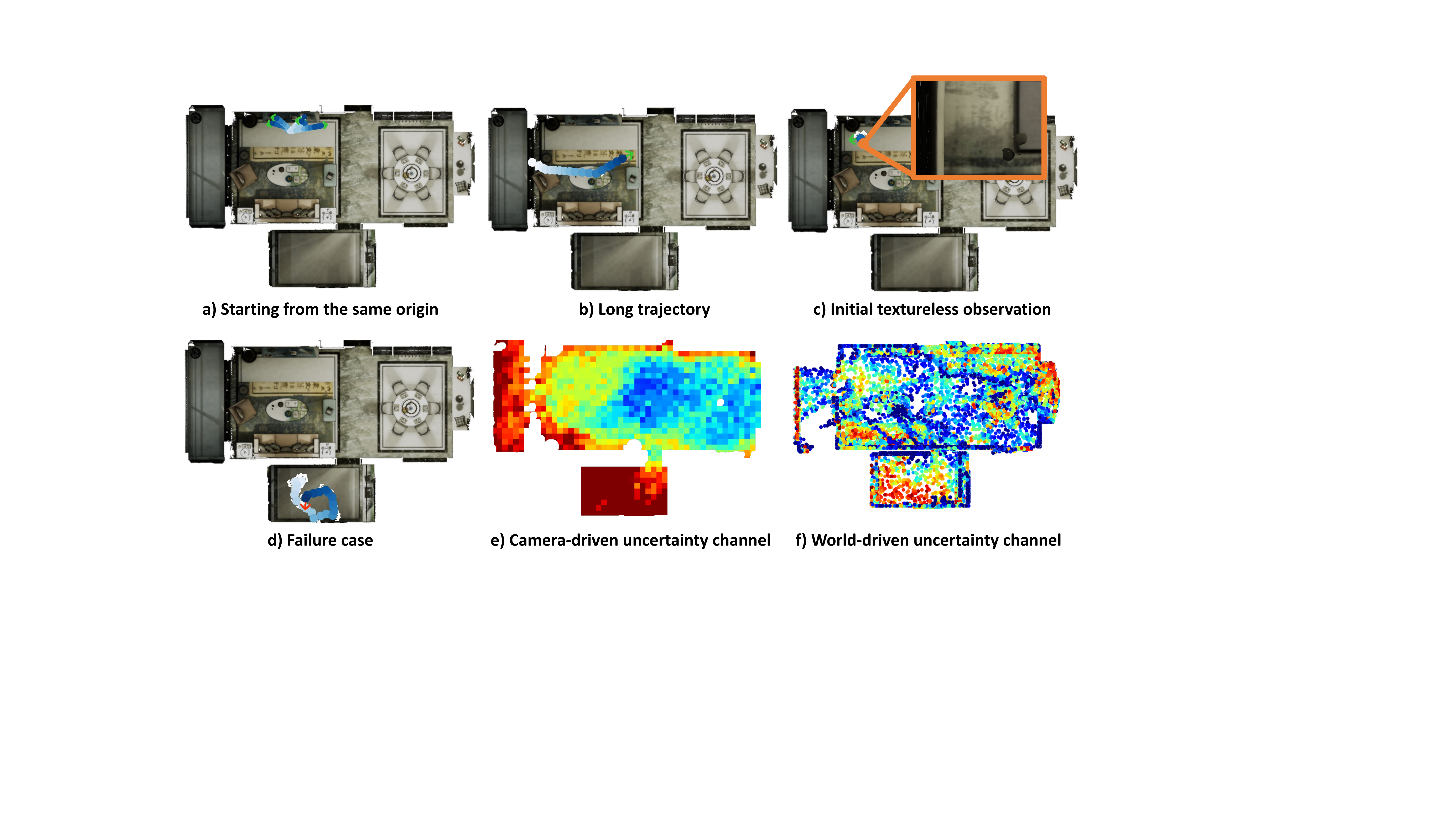}
	\end{tabular}
    \caption{Qualitative results. White arrow: start position; Green arrow: end position (successfully localized); The dots with color gradient indicate the path the agent takes. 
    Intelligent behaviors: 
    \textbf{a)} Starting from the same location, the agent travels to various regions for localization. 
    \textbf{b)} The agent is able to travel along a long trajectory for accurate localization.
    \textbf{c)} Initialized with a textureless image, the agent emerges the turn around behavior for localization.
    Failure case:
    \textbf{d)} The agent fail to get out of a small room.
    Uncertainty visualization:
    \textbf{e)} The camera-driven uncertainty channel.
    \textbf{f)} The world-driven uncertainty channel.
}
	\label{figure:visualization}
\end{figure}

\textbf{Time analysis and intelligent behavior:} It takes only 9.59s to adapt the passive localizer in a novel scene, and 0.87s to evaluate our entire algorithm for a single step, 
where the bottleneck comes from the CPU-based  implementation of ICP \cite{zhou2018open3d} (0.59s), which can be further improved with more efficient GPU implementation.
Our learned intelligent behaviors are visualized in Figure \ref{figure:visualization}.

\setlength{\tabcolsep}{2pt}	

\section{Conclusion}
In this paper, we propose a novel active camera localization algorithm, consisting of a passive and an active localization module. The former one estimates the accurate camera pose in the continuous pose space. The latter one learns a reinforcement learning policy from the explicitly modeled camera and scene uncertainty component for accurate camera localization.

\textbf{Limitation and future work:}
Figure \ref{figure:visualization} e) demonstrates a failure case, where the agent is initialized in a room with a small exit and large scene uncertainties. It fails to leave the room before reaching the maximum step length. Although we already employ a naive exploration reward to avoid repeated traversal in the same region, a smarter design, such as frontier-based exploration \cite{dornhege2013frontier} and long-term goal planning \cite{chaplot2020learning}, can be incorporated in the future for further improvement.

\textbf{Acknowledgments} 
We thank the anonymous reviewers for the insightful feedback. 
This work was supported in part by NSFC Projects of International Cooperation and Exchanges (62161146002), NSF grant IIS-1763268, a Vannevar Bush Faculty Fellowship, and a gift from the Amazon Research Awards program.

%
%
\bibliographystyle{splncs04}
\bibliography{egbib}

\appendix

\section*{Appendix}

The appendix provides the additional supplemental material that cannot be included in the main paper due to its page limit:
\begin{itemize}
\itemsep-0.0em
\item Algorithm illustration
\item More results.
\item More analysis.
\item More implementation details.
\item More details of the ACL-synthetic/real datasets.

\end{itemize}

\section{Algorithm illustration}

We summarize our proposed algorithm in Algorithm \ref{algo:main}.

\begin{algorithm*}
    \scriptsize
    \caption{The full pipeline of our algorithm}
    \label{algo:main}
    
    \begin{algorithmic}[0] 
    
        \Function{Passive Loc. Module}{observation $I^{(t)}$, posed RGB-D stream  $\big\{I_{basis}^{(i)}, C_{basis}^{(i)}\big\}_{i=1}^{m}$} 
            \If{\textit{initialization}} 
                \State \textit{initialization} $\leftarrow$ false
                \State Adapt the passive localizer by posed RGB-D stream $\big\{I_{basis}^{(i)}, C_{basis}^{(i)}\big\}_{i=1}^{m}$    
                \State Construct the scene model $D_{scene}$ by fusing posed RGB-D stream $\big\{I_{basis}^{(i)}, C_{basis}^{(i)}\big\}_{i=1}^{m}$
            \EndIf
            \State Current pose estimation $\widehat{C}^{(t)} \leftarrow$ Passive localizer$(I^{(t)})$
            \State \textbf{return} $\widehat{C}^{(t)}$
        \EndFunction
        
        \Function{Active Loc. Module}{pose estimation $\widehat{C}^{(t)}$, scene model $D_{scene}$}
            
            \State $M_{wd}^{(t)}, M_{cd}^{(t)} \leftarrow$ Scene uncertainty computation($\{\widehat{C}^{(t)}, D_{scene}\}$) 
            \State $U_{cu}^{(t)} \leftarrow$ Camera uncertainty computation($\{\widehat{C}^{(t)}, D_{scene}\}$)  
            \State Action $a^{(t)} \leftarrow$ Policy network$(\{M_{wd}^{(t)}, M_{cd}^{(t)}\})$ 
            \State \textbf{return} $U_{cu}^{(t)}$, $a^{(t)}$
        \EndFunction
        
        \Procedure{Entire Pipeline}{posed RGB-D stream $\big\{I_{basis}^{(i)}, C_{basis}^{(i)}\big\}_{i=1}^{m}$, accuracy threshold $\lambda_{cu}$}
            \State $t\leftarrow 0$
            \State $D_{scene}\leftarrow NULL$
            \State \textit{initialization} $\leftarrow$ true
            \While{$t<$ maximum step length}
                \State Obtain the current observation $I^{(t)}$
                \State $\widehat{C}^{(t)} \leftarrow$ {\scshape Passive Loc. Module}$(I^{(t)}, \big\{I_{basis}^{(i)}, C_{basis}^{(i)}\big\}_{i=1}^{m})$
                \State $U_{cu}^{(t)}$, $a^{(t)} \leftarrow$ {\scshape Active Loc. Module}$(\widehat{C}^{(t)}, D_{scene})$
                \If{$U_{cu}^{(t)}$ is within $\lambda_{cu}$ cm, $\lambda_{cu}$ degrees}
                    \State \textbf{break}
                \EndIf
                \State Execute the action $a^{(t)}$
                \State $t\leftarrow t+1$
            \EndWhile
            \State \textbf{return} $\widehat{C}^{(t)}$
        \EndProcedure
        
    \end{algorithmic}
\end{algorithm*}

\section{More results}

\subsection{Comparison on the sparse data}
The posed RGB-D stream is the basis for both passive and active localization. To further validate the robustness of our proposed algorithm, we discard half of the posed RGB-D stream as the sparse data for evaluation. The numerical comparisons with the best baselines (Camera-descent/Scene-descent) on both the sparse data and dense data (default setting in the main paper) are shown in Table \ref{table:sparse_data}. 
We observe that all the methods achieve worse results on the sparse data as expected, yet our approach still outperforms the other competitors.

\setlength{\tabcolsep}{7pt}
\begin{table}[t]
\centering
\caption{Numerical results on both the dense and sparse data.}
    \begin{tabular}{c|l|c|c|c|c}
        \toprule
        &   & \multicolumn{2}{c|}{ACL-synthetic} & \multicolumn{2}{c}{ACL-real} \\
        \cmidrule{1-6}
        Data & Methods & Acc (\%) & \#steps & Acc (\%) & \#steps \\
        \midrule
        
        \multirow{4}{*}{Dense}
        &  Camera-descent (t+1) & 61.55 & 22.90 & 61.40 & 26.85\\
        &  Camera-descent (t+2) & 55.30 & 22.60 & {59.20} & {25.78}\\
        &  Scene-descent & 57.65 & 18.56 & 54.20 & 16.87\\
        \cmidrule{2-6}
        &  Ours         & \textbf{83.05} & {17.33} & \textbf{82.40} & {17.90}\\
        
        \midrule
        \multirow{4}{*}{Sparse}  &  Camera-descent (t+1)    & 55.45 & 24.17 & 49.60 & 31.62\\
        &  Camera-descent (t+2)    & 55.05 & 28.15 & 52.40 & 37.35\\
        &  Scene-descent           & 19.90 & 6.12 & 41.40 & 22.71\\
        \cmidrule{2-6}
        &  Ours                    & \textbf{82.00} & 20.52 & \textbf{76.40} & 22.54\\
        \bottomrule
    \end{tabular}
    \label{table:sparse_data}
\end{table}

\subsection{Comparison on more real-world datasets}

To further evaluate the compatibility of our method, we compare our approach and its best competitors on 10 scenes of the real-world Gibson V2 \cite{xia2018gibson} and Replica \cite{straub2019replica} datasets besides ACL-synthetic/-real datasets in the main paper. Shown in Table \ref{table:more_data}, our results consistently outperforms the others.

\setlength{\tabcolsep}{11pt}
\begin{table}[h]
    \centering
    \caption{Numerical results on GibsonV2 and Replica datasets.}
    \begin{tabular}{l|c|c|c|c}
        \toprule
        & \multicolumn{2}{c|}{GibsonV2} & \multicolumn{2}{c}{Replica} \\
        \midrule
        Methods & Acc (\%) & \#steps & Acc (\%) & \#steps \\
        \midrule
        Camera-descent (t+1) & 57.60 & 23.51 & 67.80 & 19.04 \\
        Camera-descent (t+2) & 51.60 & 25.42 & 69.80 & 26.13  \\
        Scene-descent        & 56.20 & 16.16 & 62.80 & 14.60 \\
        \hline
        Ours         & \textbf{75.00} & 15.27 & \textbf{86.80} & 13.30 \\
        \bottomrule
    \end{tabular}
    \label{table:more_data}
\end{table}

\subsection{More qualitative results}

\begin{figure*}[t]
\centering
    \includegraphics[height=5.5cm]{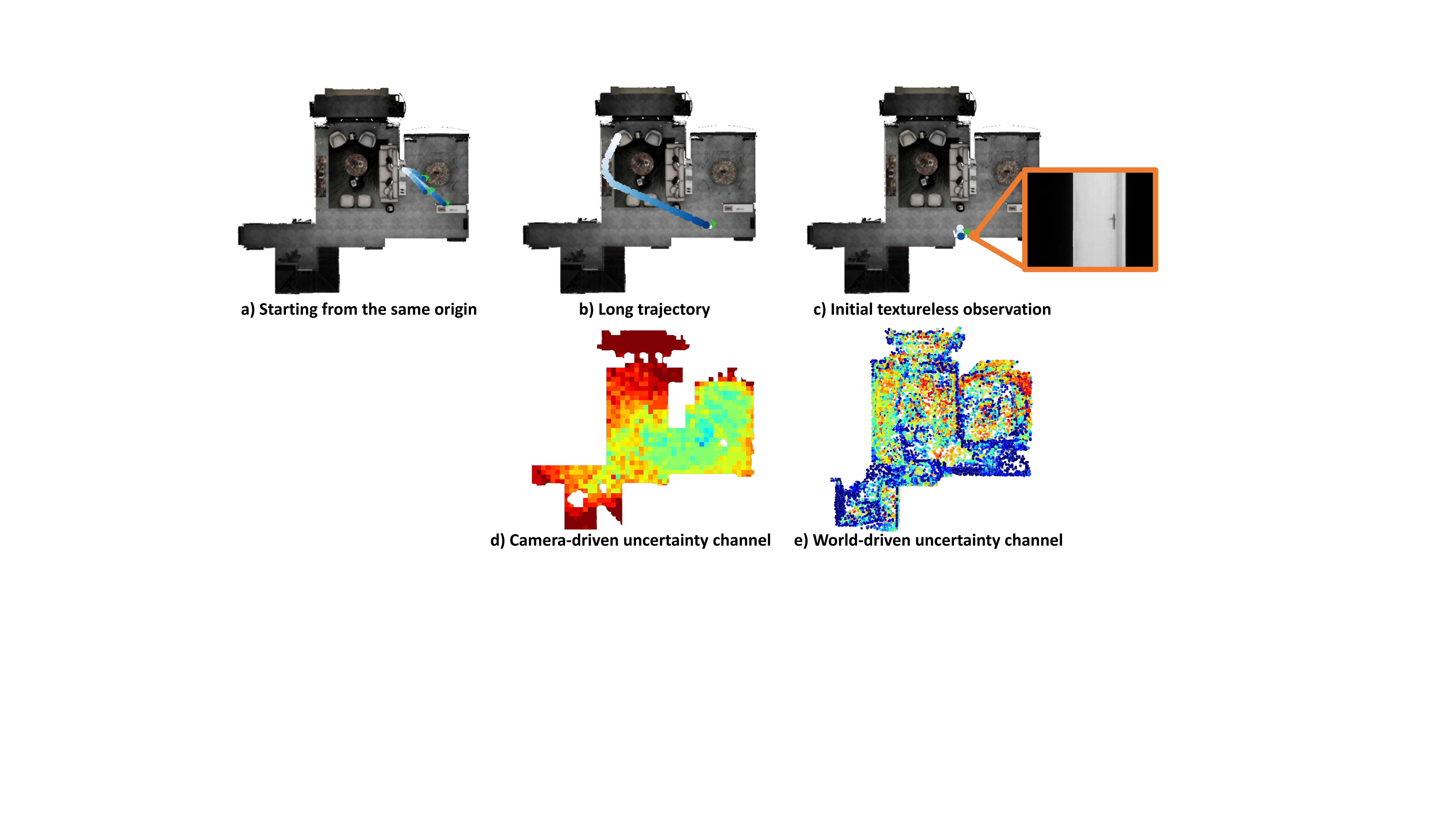} \\
    \includegraphics[height=5.3cm]{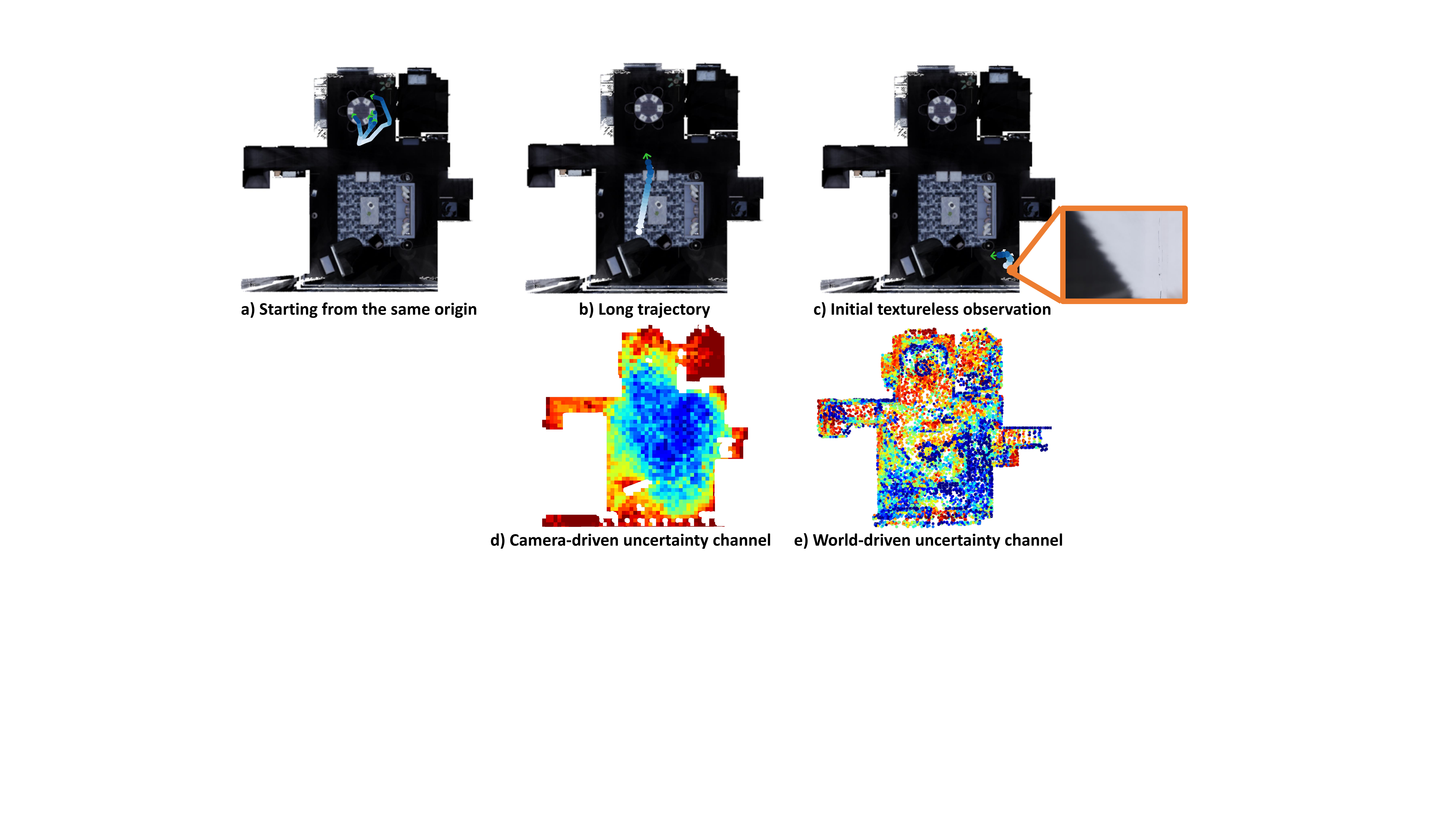} \\
    \includegraphics[height=5cm]{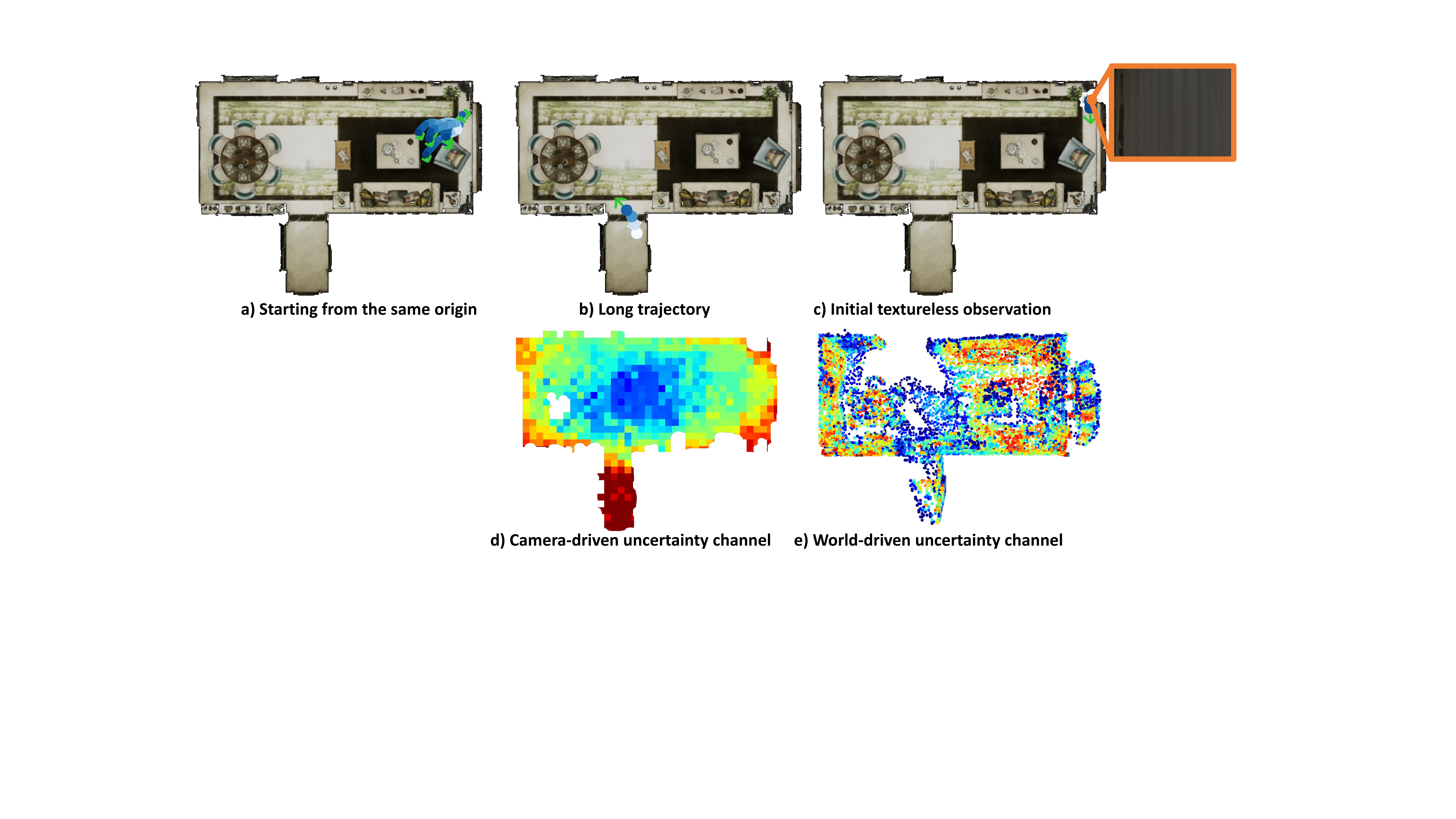} \\
	\caption{Qualitative results of the intelligent behaviors learned by our algorithm.}
	\label{figure:visual1}
\end{figure*}

In Figure \ref{figure:visual1}, we show the qualitative results of the intelligent behaviors of our algorithm on more test scenes.

\section{More analysis}

\setlength{\tabcolsep}{4pt}
\begin{table*}[t]
\centering
    \caption{Numerical results of our algorithm trained with different parameter values ($\lambda_{cu} = 5,20$) on the ACL-synthetic dataset.}
    \begin{tabular}{c|c}
        \toprule
        Uncertainty parameters & Accuracy (20cm, 20$^\circ$)\\
        \midrule
        $\lambda_{cu} = 5$ & 85.92\\
        $\lambda_{cu} = 20$ & 49.40\\
        \bottomrule
    \end{tabular}
\label{table:self_comparison_accuracy}
\end{table*}

We provide more analysis of the camera uncertainty component below.

The iterative closest point (ICP) approach is based on the general assumption that the two input point clouds are roughly aligned. When the estimated camera pose of the current frame is far from its ground truth, such as 20cm, 20$^\circ$, the camera uncertainty component generated by ICP becomes unstable and not reliable to determine the adaptive stop condition. To be specific, following the experiment of ``Analysis of camera uncertainty'' in the main paper, we further summarize that 
when the estimated relative pose is within 20cm, 20$^{\circ}$ ($\lambda_{cu} = 20$), about 83.57\% (3220/3853) samples are truly within 20cm, 20$^{\circ}$ compared to the ground truth, which is much smaller than 94.14\% for 5cm, 5$^{\circ}$  ($\lambda_{cu} = 5$).

Therefore, a natural question to ask is, when evaluating the camera pose in a coarse level, such as 20cm, 20$^\circ$, what is the best parameter value ($\lambda_{cu}$) to determine the adaptive stop condition for the highest camera pose accuracy? In Table \ref{table:self_comparison_accuracy}, we compare the numerical results of our algorithm trained with different parameter values ($\lambda_{cu} = 5/20$) and evaluated on the coarse-scale accuracy (20cm, 20$^\circ$). 
We observe that the camera pose accuracy is much worse with $\lambda_{cu} = 20$, which validates the parameter selection $\lambda_{cu}$ for the camera uncertainty component.

\section{More implementation details}

\subsection{Policy network}
The policy network takes the scene uncertainty component as input and generates the probability of the three actions defined by the action space.
The camera-driven scene map is represented as a 3-channel 2D map $M^{(t)}_{cd}$, which can be easily consumed by the convolution operation. We employ a convolution neural network of 6 convolution layers (32-64-128-128-256-256) and 1 linear layer (64) to extract the global feature ($\mathbb{R}^{64}$). Each convolution layer is of kernel size 3x3 and followed by a batch normalization layer and a max pooling layer of stride 2.
The world-driven scene map is represented as a 6-channel point cloud $M^{(t)}_{wd}$. Inspired by the popular point cloud processing network PointNet \cite{qi2017pointnet}, we employ a three-layer pointwise MLP (64-128-64) followed by a max pooling layer to extract its global feature ($\mathbb{R}^{64}$).
Finally, by concatenating all the extracted features, we use a three-layer MLP (64-16-3) to predict the probability of the three predefined actions.

\subsection{Noise perturbation on the action space}

To simulate robotic agents in a real-world condition, the action does not lead to perfect execution, hence we add the Gaussian noise to each action. To be specific, if the agent turns left or right, the Gaussian noise of standard deviation (6) will be added to the rotation angle $\theta^{(t)}$ of mean value (20); if the agent moves forward, the Gaussian noise of standard deviation (5) will be added on the 2D positions $x^{(t)},y^{(t)}$ of mean value (30). The positions are measured in centimeters, and the rotation angle is measured in degrees. Note the standard deviation ($\sigma$) is actually very large considering 31.74\% of sampled noises are beyond $\sigma$ for the Gaussian distribution.

\subsection{Implimentation details}
In our experiment, we employ the Adam \cite{kingma2014adam} to optimize the network weights with the initial learning rate of $3\times 10^{-4}$. Some hyper-parameters: $N_{cd} = 12, N_{wd\_r} = 1000, N_{wd\_p} = 2^{14} = 16384, N_f = 5, X = 256, Y = 256$.

\section{More details of the ACL-synthetic/real datasets}

The posed RGB-D stream in the existing camera localization datasets \cite{shotton2013scene,valentin2016learning,wald2020beyond} is usually obtained by scanning the environment with handheld sensors by human operators, hence does not always cover the complete scene model. We design the posed RGB-D stream in our dataset to simulate this effect.
Directly visualizing the trajectory of the posed RGB-D stream in the scene is not intuitive as it would lose the orientation information of the camera pose, instead we choose to visualize the scene model reconstructed from the posed RGB-D stream to showcase how much scene region is covered by the posed RGB-D stream. 

\begin{figure*}[t]
\centering
    \includegraphics[width=12cm]{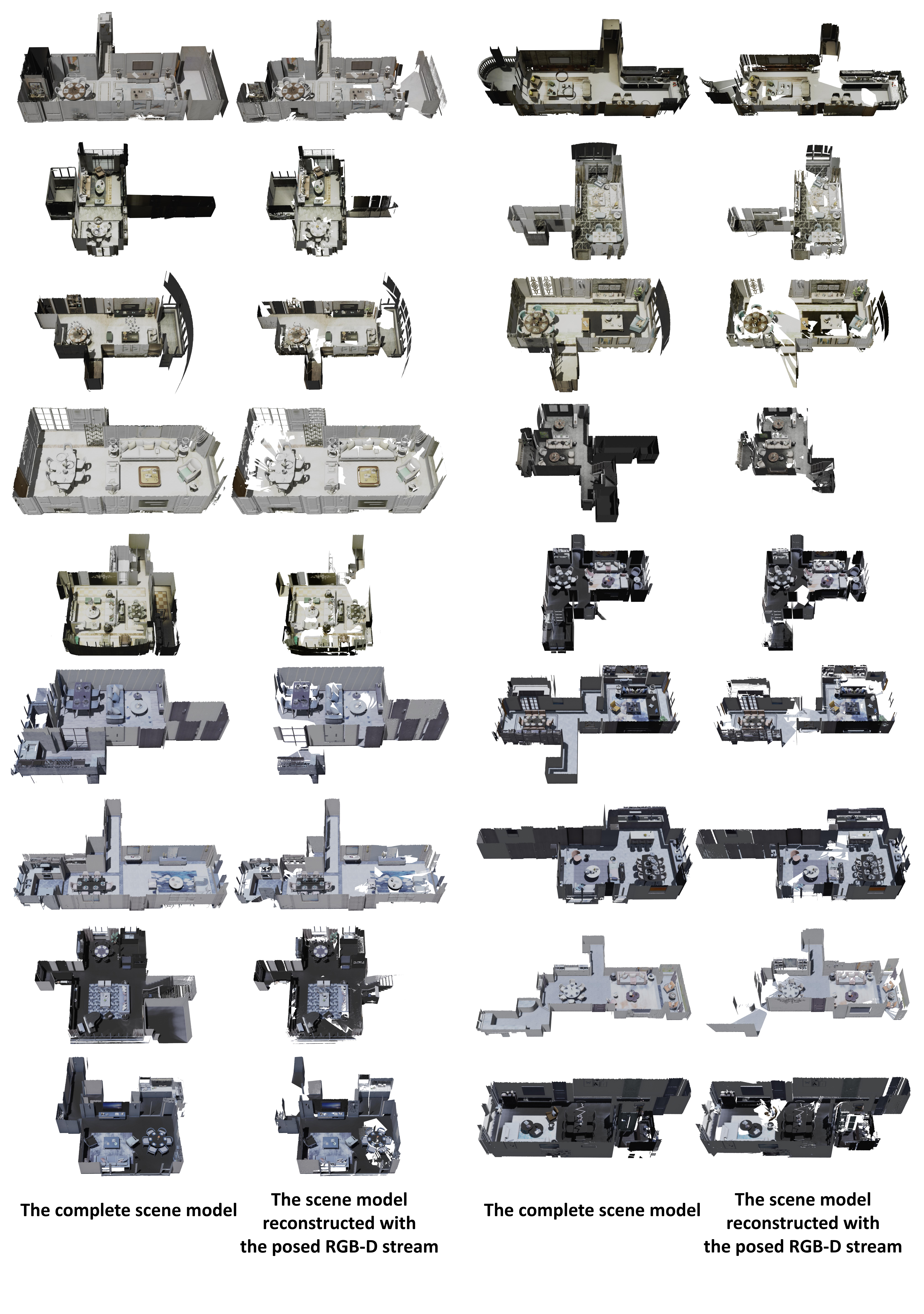}
	\caption{Visualization of the ACL-synthetic dataset.}
	\label{figure:acl_syn1}
\end{figure*}

\begin{figure*}[t]
\centering
    \includegraphics[width=12cm]{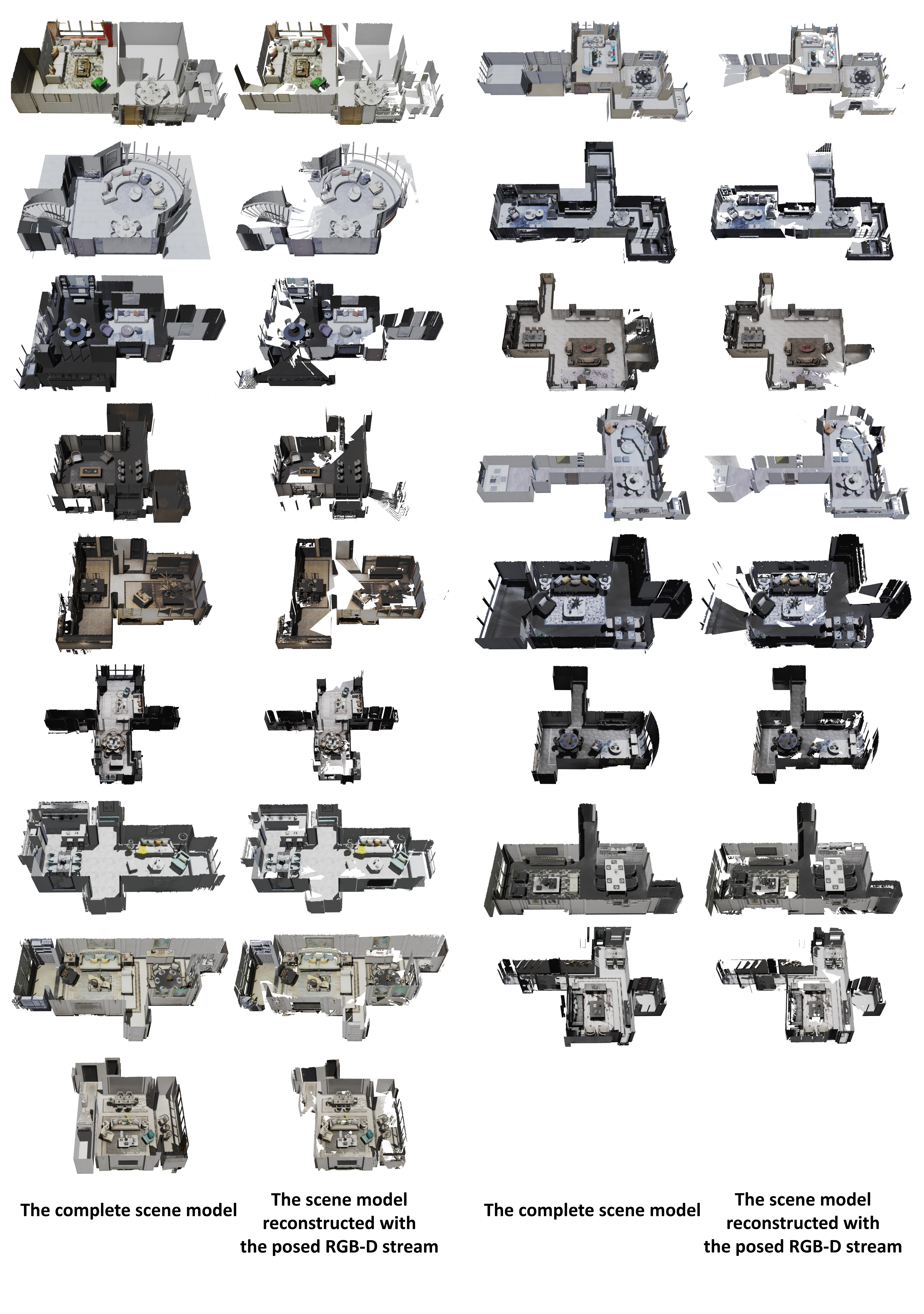}
	\caption{Visualization of the ACL-synthetic dataset.}
	\label{figure:acl_syn2}
\end{figure*}

\begin{figure*}[t]
\centering
    \includegraphics[width=12cm]{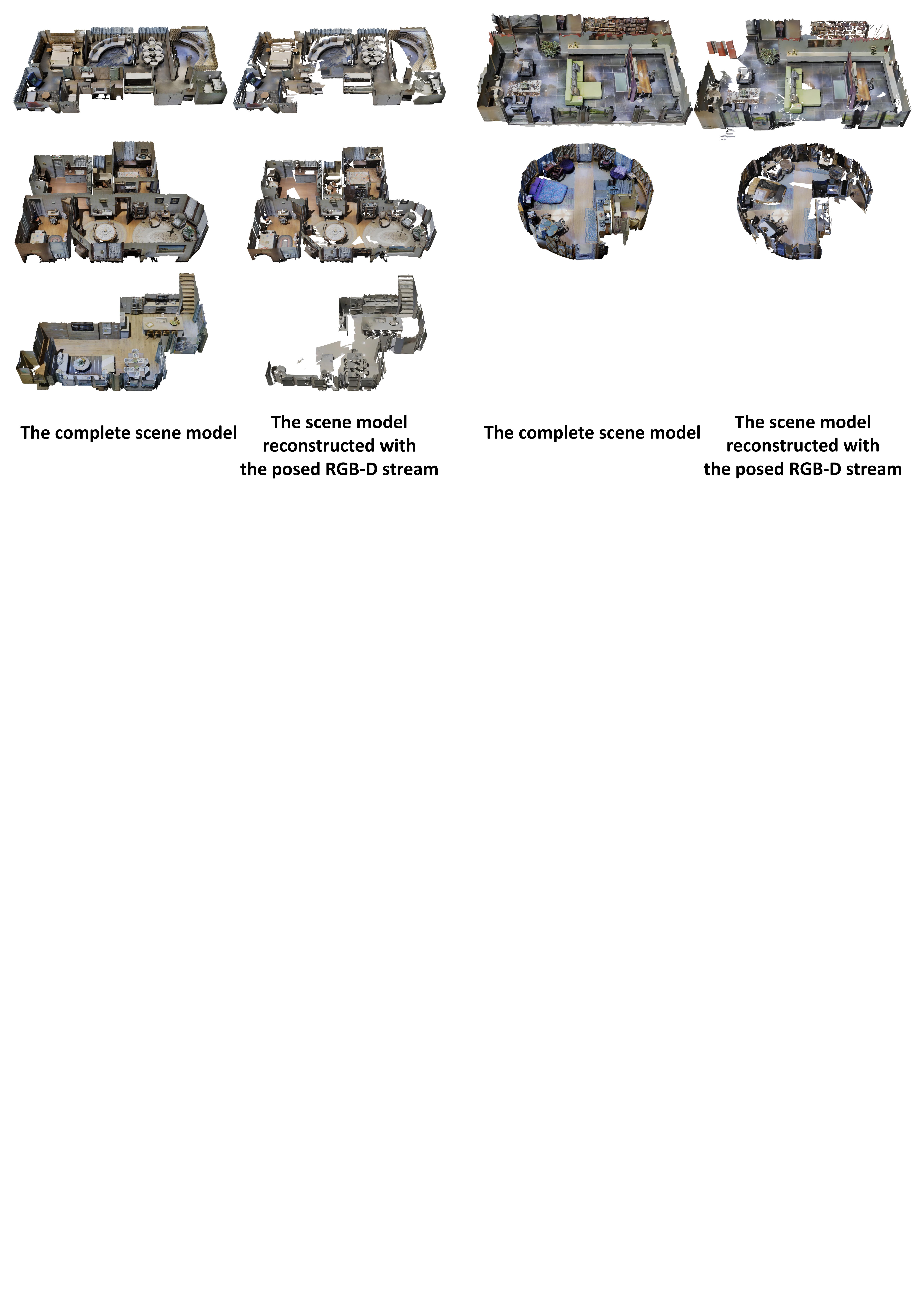}
	\caption{Visualization of the ACL-real dataset.}
	\label{figure:acl_real}
\end{figure*}

We illustrate the textured meshes of both the complete and reconstructed scene models for the ACL-synthetic and ACL-real datasets in Figure \ref{figure:acl_syn1}, \ref{figure:acl_syn2} and \ref{figure:acl_real}.
Their related statistics are shown in Table \ref{table:statistics}.

\setlength{\tabcolsep}{4pt}
\begin{table}[t]
  \centering
  \caption{Scene statistics of the ACL-synthetic and ACL-real dataset. We summarize the number of scenes, scene area, max scene area, min scene area and the number of frames in the RGB-D sequences. The unit for all areas is $m^2$. The Area and \#frames metrics are averaged over all the scenes involved.}
    \begin{tabular}{llccccc}
    \toprule
    \multicolumn{2}{l}{Scene} & \#scenes & Area & Max area & Min area & \#frames \\
    \midrule
    \multirow{2}{*}{ACL-synthetic} & Train split &       15&       37.89 & 49.40 & 25 & 58.00  \\
                              & Test split &       20&       43.17 & 75.00  & 26.9 & 54.45  \\
    \midrule
    ACL-real & Test split &       5&    64.82  &  98.28 &  23.62  &  88.40\\
    \midrule
    \multicolumn{2}{l}{All} &       40&       43.90 &  98.28  &  23.62 &60.03  \\
    \bottomrule
    \end{tabular}
  \label{table:statistics}
\end{table}

\end{document}